# Deployable polyhedrons with one-DOF radial transformation

Yuanqing Gu [1], Yan Chen [1,2,*]

[1] *School of Mechanical Engineering, Tianjin University, Tianjin 300350, China*
[2] *Key Laboratory of Mechanism Theory and Equipment Design of Ministry of Education, Tianjin University, Tianjin 300350, China*

**Abstract**

**Deployable polyhedrons can transform between Platonic and Archimedean polyhedrons to meet the demands of various engineering applications. However, the existing design solutions are often with multiple degrees of freedom and complicated mechanism links and joints, which greatly limited their potential in practice. Combining the fundamentals of solid geometry and mechanism kinematics, this paper proposes a family of kirigami Archimedean polyhedrons based on the *N*-fold-symmetric loops of spatial 7*R* linkage, which perform one-DOF radial transformation following tetrahedral, octahedral, or icosahedral symmetry. Moreover, in each symmetric polyhedral group, three different transforming paths can be achieved from one identical deployed configuration. We also demonstrated that such design strategy can be readily applied to polyhedral tessellation. This work provides a family of rich solutions for deployable polyhedrons to facilitate their applications in aerospace exploration, architecture, metamaterials and so on.**

* Corresponding author. *E-mail address*: yan_chen@tju.edu.cn (Y. Chen)

# Introduction

Deployable hollow polyhedrons enable large volumetric change that fascinates both mathematicians and engineers[1-2]. The transformable polyhedrons, a special type of deployable polyhedrons, can realise the shape transformation between two regular polyhedral configurations, which are mathematically interesting yet kinematically challenging[3-4]. Buckminster Fuller first provided a solution known as the jitterbug[5], in which the transformation from a cuboctahedron to an octahedron was identified. The proposed cuboctahedron has only interlinked triangular rigid faces without any square faces, and during the transformation, the triangle faces rotate with respect to each other until they are joined together edge-by-edge, leading to an octahedron[6]. This is kinematically equivalent to a mechanism system with six degrees of freedom (DOFs), in which the neighbouring triangles are connected by spherical joints at the polyhedral vertices, thus the transformation will be completed in an orderly way. Nevertheless, the elegant motion is possible only under the symmetry constraints, i.e., the triangles translate and rotate along their virtual central axis[7].

Several similar solutions to this classic problem have also been proposed, yet most contain multiple DOFs, meaning that certain means of symmetry constraints must be imposed in the transformation process. Moreover, inspired by the inventions of Buckminster Fuller, Verheyen has reported various Platonic polyhedral linkages, yet also with multiple DOFs and unorderly motion due to the spherical joints used at the polyhedral vertices[8]. Based on the expandable dodecahedron, Kovács et al.[9-10] proposed a class of multiple-DOF expandable polyhedral structures by using 2-DOF joints at each vertices of pentagonal panels. It is challenging to conduct the transformation in an orderly way through mechanical coupling.

To solve the control problem and fold a polyhedron in unison, one-DOF transformation of the deployable polyhedral mechanism (DPM) is desirable without breaking the original polygonal panels[11-12]. For instance, Yang et al.[13-15] accomplished three paired one-DOF solutions between Archimedean and Platonic polyhedrons by means of spatial overconstrained linkages and large motion-range spherical joints, while the polyhedral symmetry is broken in the folding process. Wohlhart[16] proposed a variety of twisting polyhedrons by setting up gussets at the vertices, in which one-DOF transformation between two Archimedean polyhedrons can be obtained also with an asymmetric twisting motion. Similarly, various transformable polyhedrons have been constructed to mechanically transform the cyclic polyhedrons into their corresponding dual forms[17-20]. However, those asymmetric polyhedral mechanisms lose the original symmetries in their transformations, which limits their potential for 3D tessellation in engineering applications such as space modular equipment and metamaterials[21].

On the other hand, some one-DOF polyhedral mechanisms can preserve the original symmetry during the entire motion process due to the radial motion[22-24], but they cannot realize regular configurational transformations between two different polyhedrons[25-27]. There is little work on the construction of DPMs together with one-DOF radial motion and rich configurational transformations. Nevertheless, our previous work proposed a group of radial origami polyhedrons with symmetric transformability[28], yet only hexagon facets of polyhedrons can be foldable resulting in the limited capability in rich transformations.

In this paper, we propose a family of kirigami Archimedean polyhedrons based on spatial 7$R$ linkages and their $N$-fold-symmetric loops, which perform one-DOF synchronized radial motion and symmetric transformability following $T_d$, $O_h$ and $I_h$ symmetries, respectively. In each symmetric polyhedral group, various transformations can be achieved from one identical deployed configuration to three different folded configurations, as well as the transformable solutions among Platonic and Archimedean polyhedrons.

## Results

**Construction and transformation of kirigami Archimedean polyhedrons.**

For the convex regular polyhedrons in 3D Euclidean space, their faces are congruent regular polygons and are assembled in the same way around each vertex, in which two classic groups are Platonic polyhedrons and Archimedean polyhedrons[29]. Most of them can be classified according to three different symmetries, i.e., tetrahedral symmetry ($T_d$), octahedral symmetry ($O_h$) and icosahedral symmetry ($I_h$) [30]. Meanwhile, some interesting polyhedral pairs can be identified through mathematical transformation principle (Supplementary Note 1), in which each pair of polyhedrons has the same polyhedral symmetry due to their radial transformation in geometry.

Using kinematic strategies to realise such polyhedral transformations obtained in geometry, the first task is to fold polygons on the polyhedral surface. For instance, to transform a truncated tetratetrahedron into a truncated tetrahedron, as illustrated in highlighted red dotted line area in Fig. 1a, the yellow hexagon facet should be folded into a triangle as well as the adjacent cyan squares are vanished, while other blue hexagon facets move synchronously and radially with respect to the polyhedral centroid. For this purpose, see Fig. 1b, to completely fold the cyan squares, the design strategy of the construction unit is originated from a kirigami pattern with nine sheets ($p_1$ to $p_9$) when we cut the yellow hexagon facet and add an additional valley crease on each cyan square[26]. It can be kinematically modelled as a threefold-symmetric spatial $9R$ linkage (a linkage with nine revolute joints) that can be divided into three sets of parallel revolute joints, i.e., $z_2 // z_3 // z_4$, $z_5 // z_6 // z_7$ and $z_8 // z_9 // z_1$, and other design parameters are depended on polyhedral geometry, yet it has three DOFs. To obtain one-DOF folding motion, extra motion constraints need to be introduced into this kirigami pattern. Here, we add four additional yellow sheets ($q_1$ to $q_4$) to further associate and constrain the motion of blue platforms ($p_1$, $p_4$ and $p_7$), in which $q_1$ to $q_3$ are identical isosceles trapezoid and $q_4$ is an equilateral triangle. To match the threefold symmetry, sheet $q_4$ should be set up on the center of original yellow hexagon, and each sheet of $q_1$ to $q_3$ connects the central sheet $q_4$ and the corresponding blue platforms with two parallel joints. As a result, a modified kirigami pattern is obtained (Fig. 1b) and can perform a one-DOF threefold-symmetric synchronized folding motion (kinematic analysis and geometry conditions can be found in Supplementary Note 2), which can be modelled as a threefold-symmetric 7$R$ loop that consists of three 7$R$ linkages $p_1p_2p_3p_4q_2q_4q_1$, $p_4p_5p_6p_7q_3q_4q_2$ and $p_7p_8p_9p_1q_1q_4q_3$, where two adjacent 7$R$ linkages share three common panels and two related parallel joints.

Regarding a construction unit, see Fig. 1c, the symmetric 7$R$ loop can be tessellated with tetrahedral symmetry ($T_d$) to synthesize kirigami truncated tetratetrahedron polyhedron and realise the transformation as given in Fig. 1a. First, a quarter portion of the entire polyhedral surface is symmetrically highlighted by gray dotted lines according to $T_d$ symmetry. Then, we embed one threefold-symmetric 7$R$ loop into this quarter polyhedral surface referring to polyhedral geometry and $T_d$ symmetry. Finally, the kirigami truncated tetratetrahedron is constructed by embedding four 7$R$ loops following $T_d$ tessellation. The one-DOF construction unit and their symmetric tessellation guarantee that the kirigami truncated tetratetrahedron also performs a one-DOF synchronized folding motion (the detailed kinematic proof is given in Supplementary Note 2), whose cardboard prototype is shown in Fig. 1c. Therefore, the polyhedral transformation from a truncated tetratetrahedron (left) to a truncated tetrahedron (right) is ultimately accomplished during the entire folding process.

Furthermore, if we fold a truncated cuboctahedron into a truncated octahedron with $O_h$ symmetry, the corresponding octagons need to be folded into squares based on the transformation principle, in which the threefold-symmetric 7$R$ loop is not applicable. To solve this problem, the similar one-DOF fourfold-symmetric 7$R$ loops are introduced, see Fig. 1d, which consists of four spatial 7$R$ linkages at four hollows. Two adjacent 7$R$ linkages share two related parallel joints on the corresponding blue trapezoid to guarantee one-DOF synchronized folding motion of this construction unit. We embed one fourfold-symmetric 7$R$ loop into the one-sixth polyhedral surface, then following $O_h$ tessellation, a kirigami truncated cuboctahedron is obtained to fulfill the corresponding polyhedral transformation. Similarly, we can also use the one-DOF fivefold-symmetric 7$R$ loops to deal with the folding of decagon and create a kirigami truncated icosidodecahedron following icosahedral symmetry ($I_h$). As illustrated in Fig. 1e, a total of twelve fivefold-symmetric 7$R$ loops are embedded into the surface of this construction, and the transformation from a truncated icosidodecahedron to a truncated icosahedron is simultaneously revealed.

So far, using the threefold-, fourfold- and fivefold-symmetric loops to respectively fold the hexagon, octagon and decagon on different polyhedral surfaces, three types of kirigami Archimedean polyhedrons with various symmetries are obtained, as well as the corresponding one-DOF radial polyhedral transformations. In the above polyhedral symmetric constructions, it can be found that there must be one type of polygon will be reserved and perform translating motion with respect to the polyhedral centroid, while the facets in other colors are folded. If we separately select different types (colors) of polygons in one polyhedron to fold or reserve, a richer variety of polyhedral transformations can be realized by means of the proposed $N$-fold-symmetric $7R$-loop construction unit.

**Polyhedral transformations with different folding paths.** Take a truncated tetratetrahedron in Fig. 2a as an example in which these types of polygons are illustrated in different colors, we can select to reserve one type of polygons (marked in red line) and fold the other two, respectively, leading to three possible transformations in solid geometry along three different transformation paths (TP 1.1 to TP 1.3). First, we reserve the blue hexagons in TP 1.1, and fold the other types by using the proposed threefold-symmetric $7R$ loops, whose corresponding kinematic solution has been demonstrated in Fig. 1c. Second, due to the duality of the tetrahedral group, the construction by reserving yellow hexagons in TP 1.2 also results in the same transformation as given in TP 1.1. Third, if we reserve cyan squares in TP 1.3, the synchronous folding of yellow and blue hexagons can create a different transformation from a truncated tetratetrahedron to a rhombitetratetrahedron. However, we rotate the threefold-symmetric yellow and blue sheets with 60 degrees to ensure the connection with the reserved cyan squares, respectively, leading to a one-DOF assembly of spatial $8R$ linkages instead of $7R$ linkages. Therefore, the three transformations from an identical deployed configuration to three folded configurations are revealed, see Supplementary Movie 1.

Furthermore, for the Archimedean polyhedrons with $O_h$ symmetry, as illustrated in Fig. 2b, by reserving the blue octagon (marked in red line) and using a threefold-symmetric $7R$ loop as the construction unit to fold the yellow hexagons and cyan squares, the transformation from truncated cuboctahedron to truncated cube is firstly obtained in TP 2.1. Next, the transformation in TP 2.2 has been presented by introducing six fourfold-symmetric 7*R* loops to fold the blue octagons, also see Fig. 1d. Finally, by combining the above folding of yellow hexagons and blue octagons whilst reserving cyan squares, the transformation to rhombicuboctahedron is obtained, which also results in an assembly of 8*R* linkages. Here, three different folded configurations are revealed (Supplementary Movie 2), and the one-DOF synchronized radial motion with $O_h$ symmetry is always presented in each transformation of kirigami polyhedrons in Fig. 2b.

Moreover, we can readily create a series of one-DOF kirigami polyhedrons with $I_h$ symmetry and their corresponding transformations, see Fig. 2c. Embedding threefold-symmetric 7*R* loops, fivefold-symmetric 7*R* loops and their combination in a truncated icosidodecahedron result in the transformations into a truncated dodecahedron (TP 3.1), a truncated icosahedron (TP 3.2, also see Fig. 1e) and a rhombicosidodecahedron (TP 3.3), respectively, see Supplementary Movie 3.

By now, the radial transformations among Archimedean polyhedrons are revealed following tetrahedral, octahedral and icosahedral symmetries, respectively. Here, the interesting but challenging mathematical issue has been solved with mechanism strategy, i.e., from one identical deployed configuration, different folded configurations in each symmetric polyhedral group are obtained by means of *N*-fold-symmetric construction units.

**Transformations among Archimedean and Platonic polyhedrons.** Based on the proposed kirigami polyhedrons, we can also explore the transformation solutions from Archimedean to Platonic polyhedrons by following the transformation principle. As shown in Fig. 3a, the blue triangles in a rhombitetratetrahedron are reserved in the geometric transformation, while the other facets should be folded to create a tetrahedron.

Here, to embed the threefold-symmetric 7*R* loops, we extend three edges with the width of $h$ at the vertices of each original triangle to accommodate the embedding of 7*R* loops, respectively, in which the width $h$ can theoretically be infinitely small. Thus, the kirigami rhombitetratetrahedron is obtained as well as its intermediate transformation to a tetrahedron. Note that, only carrying out the geometric variations, the kirigami rhombitetratetrahedron (Fig. 3a) has the same mechanism topology as the kirigami truncated tetratetrahedron (TP 1.1 in Fig. 2a), i.e., isomorphic assembly of threefold-symmetric 7*R* loops.

Similar to the solution as presented in Fig. 3a, we can also obtain more paired transformations among Archimedean and Platonic polyhedrons with $O_h$ and $I_h$ symmetries. As shown in Fig. 3b, a deployed rhombicuboctahedron can be transformed into a cube when only the blue square facets are reserved following the transformation principle. Thus, we use the threefold-symmetric 7*R* loops to fold the yellow triangles as well as the related cyan facets by combining the geometric variations, which consequently results in the transformation into a cube. On the other hand, to fulfill the transformation from the identical rhombicuboctahedron to an octahedron by reserving the yellow triangles, fourfold-symmetric 7*R* loops are adapted to fold the blue squares, and the corresponding result is shown in Fig. 3c. Thus, the kinematic solution as demonstrated in Fig. 3b has the same mechanism topology as the one in TP 2.1 in Fig. 2b, so do the case in Fig. 3c as the one given in TP 2.2. It can also be found that the two folded Platonic polyhedrons in Figs. 3b and 3c are dual due to the transformation principle.

Moreover, we can readily create the transformations from the identical rhombicosidodecahedron into a dodecahedron and an icosahedron with $I_h$ symmetry, respectively. Here, two kirigami rhombicosidodecahedrons are constructed by embedding threefold- and fivefold-symmetric 7R loops, as shown in Figs. 3d and 3e, in which the edges with the small width at the vertices of each reserved pentagon and triangle have been extended to accommodate the 7R loops, respectively. Thus far, the transformable solutions from five Archimedean into all five Platonic polyhedrons are revealed.

No matter which transformation in Fig. 3, the original symmetry is always reserved in the continuous folding process of each kirigami polyhedron, as well as the one-DOF synchronized radial motion. Meanwhile, in addition to two special Archimedean polyhedrons without symmetry, i.e., snub cube and snub dodecahedron, the possible one-DOF transformations among the rest of eleven Archimedean and all five Platonic polyhedrons can be obtained (details can be found in Supplementary Note 3).

**Superimposed pattern and polyhedral tessellations.** Different polyhedral mechanisms have identical deployed configurations that can provide inspiration for crease superposition. Taking the kirigami truncated cuboctahedrons with $O_h$ symmetry as an example, we can obtain a superimposed pattern with two different folding paths, see Fig. 4a. The red pattern, $K_1$, has been demonstrated in TP 2.1 in Fig. 2b with the transformation into truncated cube, and the black one, $K_2$, has been given in TP 2.2 with the transformation into truncated octahedron, both have one-DOF radial foldability. Next, patterns $K_1$ and $K_2$ are superimposed on the surface of a common truncated cuboctahedron to form the pattern $K_{1,2}$. Due to the vertical intersected creases on each cyan square, the actuation of red pattern $K_1$ will constrain all creases and vertices of pattern $K_2$, i.e., $K_2$ is held inactive [31]. Thus, pattern $K_{1,2}$ can be independently folded following TP 2.1, meaning that the folding of one motion path suppresses the other, and vice versa (Supplementary Movie 4). The superimposed pattern thus reserves the one-DOF radial folding motion of the original patterns, as shown in the prototypes following

TP 2.1 and TP 2.2, respectively. The pattern superimposition can also be utilized for the kirigami polyhedrons with $T_d$ and $I_h$ symmetries, which demonstrates the potential to enable one kirigami polyhedron with two or more folded configurations.

Meanwhile, also following the Oh symmetry, those two mentioned kirigami truncated cuboctahedrons can be tessellated in 3D space while retaining the one-DOF radial foldability. As shown in Fig. 4b, we tessellate a truncated cuboctahedron along tessellation path 1 (TeP 1) by sharing the blue octagons to obtain, for example, a 2×2×2 tessellation result[32]. Based on this 3D tessellation, we can apply the above kirigami pattern $K_1$ and $K_2$ into each unit of this tessellation along TP 2.1 and TP 2.2, respectively, thus two different folded tessellation results are obtained with synchronized foldability. In addition, more 3D tessellations with Oh symmetry along various transformation paths can be found in Supplementary Note 4.

## Discussion

In summary, we have designed a family of kirigami Archimedean polyhedrons with one-DOF synchronized radial motion based on spatial 7R linkages and their *N*-fold-symmetric loops, and the corresponding prototypes are fabricated to verify their deployable transformability and kinematic properties. Polyhedral transformations have been demonstrated following $T_d$, $O_h$ and $I_h$ symmetries respectively, in which three different transforming paths can be achieved from one identical deployed configuration in each symmetric polyhedral group. The pattern superimposition has been proposed to enable a single kirigami polyhedron with diverse folding configurations, as well as the $O_h$ tessellations with different transformation paths. The reported design methodology can be readily extended to the design of three-dimensional deployable structures such as Johnson polyhedrons. Hence, the polyhedral solutions proposed in this paper could attract the attention of both mathematicians and engineers, due to their potential applications in the fields of manufacturing, architecture and space exploration, such as human habitats on Mars that need to be folded neatly for launch.

The richer transformable solutions among Archimedean and Platonic polyhedrons have been investigated in this paper, future work will explore other possible polyhedral pairs and the corresponding kinematic solutions by considering the wider polyhedral symmetry. It should be noted that the proposed polyhedrons are multiloop mechanisms with many redundant constraints, the constraint reduction strategy could be further explored to find the simplified constraint system with appropriate stiffness. This work also brings a kinematic strategy to create mechanism-based metamaterial unit, their control approach and advanced applications in multifunctional and programmable metamaterials should be developed extensively.

## Data availability

All the data supporting the findings of this study are available from the corresponding authors upon reasonable request. Source Data are provided with this paper.

## Code availability

The computer code and algorithm that support the findings of this study are available from the corresponding author upon reasonable request.

# Acknowledgements

Y.C. acknowledged the support of the National Natural Science Foundation of China (Projects 52320105005, 52035008, 51825503) and the New Cornerstone Science Foundation through the XPLORER PRIZE (XPLORER-2020-1035).

# Author contributions

Y.C. supervised the research. Y.Q.G. generated models and analyzed the data. Y.C. and Y.Q.G. discussed the results. All authors wrote the paper.

# Competing interests

Authors declare no competing interests.

# Additional Information

**Supplementary Materials**

Supplementary Text

Supplementary Movies 1 to 4

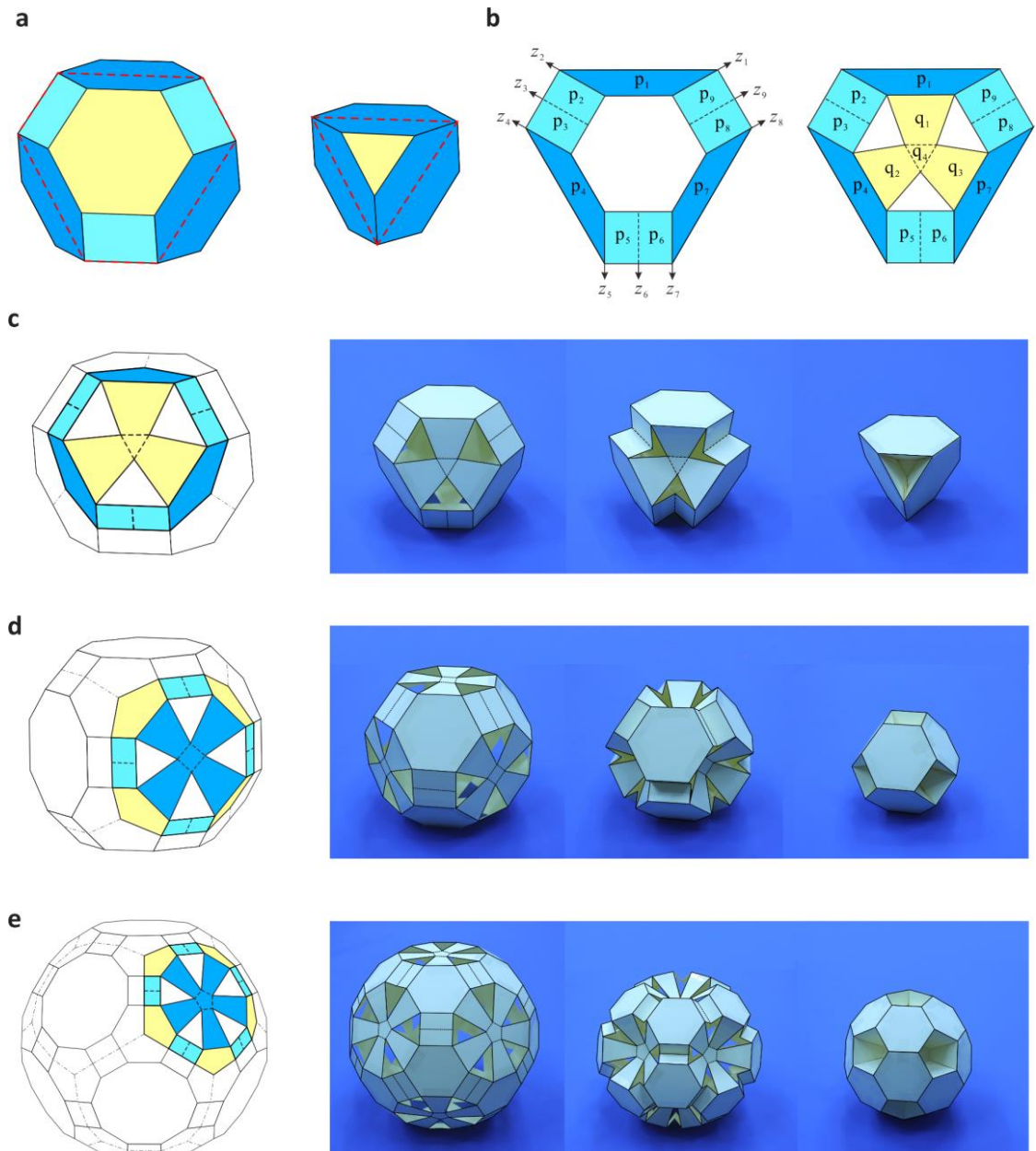

**Fig. 1 Construction of kirigami deployable polyhedrons with 7R loops. a** Transformation from a truncated tetratetrahedron to a truncated tetrahedron in geometry. **b** Threefold-symmetric 7R loop based on a spatial 9R linkage. **c** Kirigami truncated tetratetrahedron constructed by threefold-symmetric 7R loops, and the transformation sequence from a truncated tetratetrahedron to a truncated tetrahedron. **d** Kirigami truncated cuboctahedron constructed by fourfold-symmetric 7R loops, and the transformation sequence from a truncated cuboctahedron to a truncated octahedron. **e** Kirigami truncated icosidodecahedron constructed by fivefold-symmetric 7R loops, and the transformation sequence from a truncated icosidodecahedron to a truncated icosahedron.

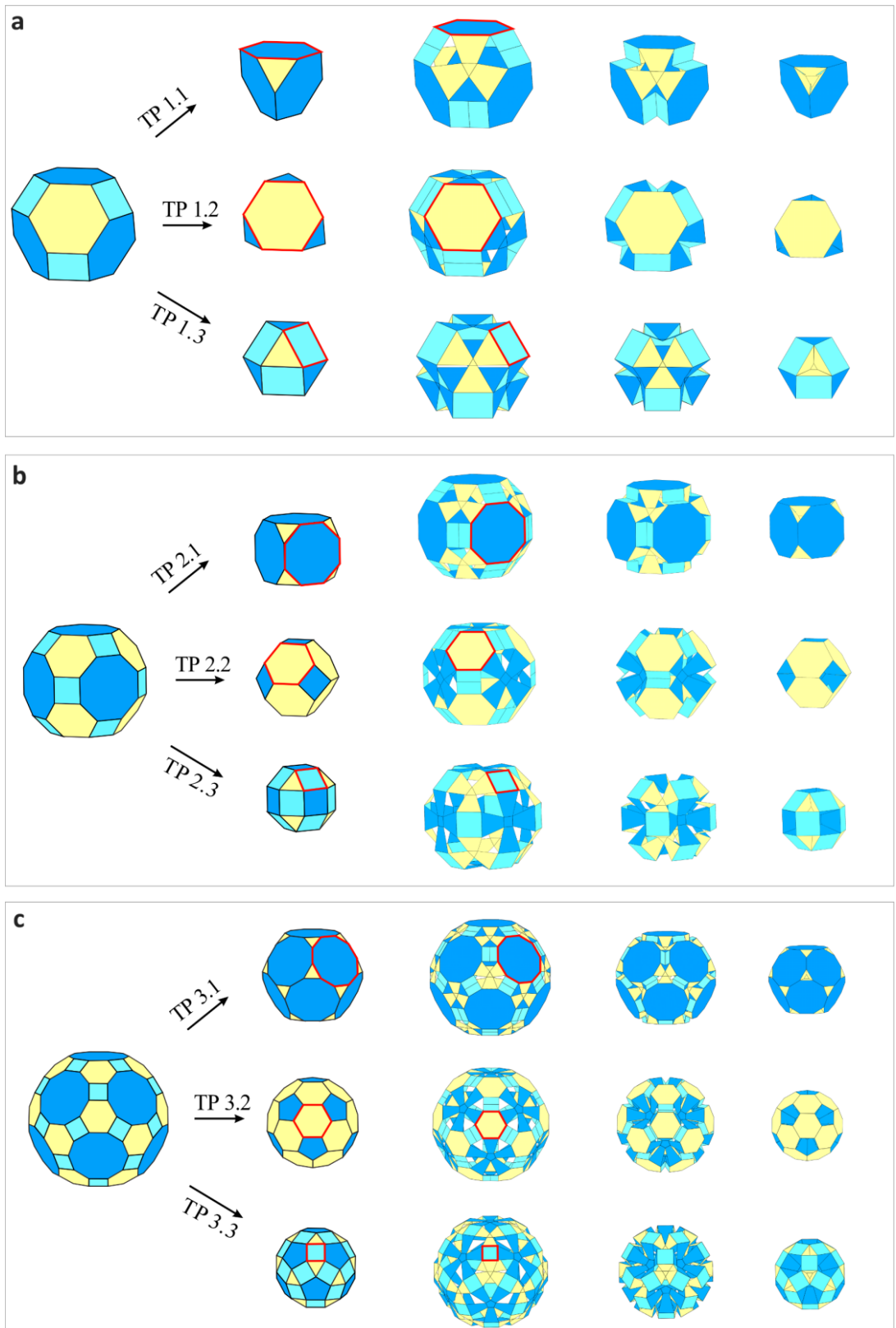

**Fig. 2 Transformations of kirigami polyhedrons with different folding paths. a** Transformations from a truncated tetratetrahedron with $T_d$ symmetry to two truncated tetrahedrons and a rhombitetratetrahedron, respectively. **b** Transformations from a truncated cuboctahedron with $O_h$ symmetry to a truncated cube, a truncated octahedron and a rhombicuboctahedron, respectively. **c** Transformations from a truncated icosidodecahedron with $I_h$ symmetry to a truncated dodecahedron, a truncated icosahedron and a rhombicosidodecahedron, respectively.

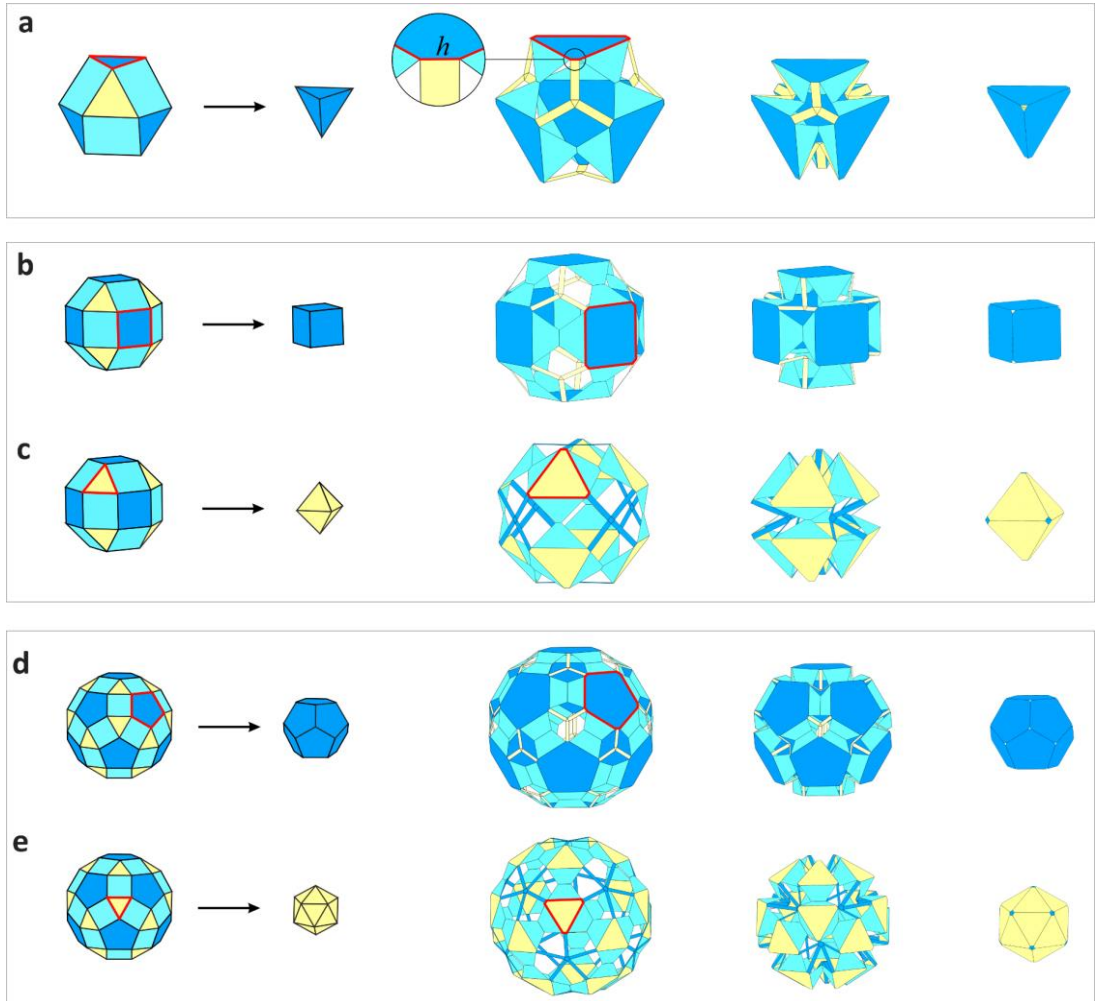

**Fig. 3 Transformations among Archimedean and Platonic polyhedrons. a** Transformation from a rhombitetratetrahedron to a tetrahedron. **b** Transformation from a rhombicuboctahedron to a cube. **c** Transformation from a rhombicuboctahedron to an octahedron. **d** Transformation from a rhombicosidodecahedron to a dodecahedron. **e** Transformation from a rhombicosidodecahedron to an icosahedron.

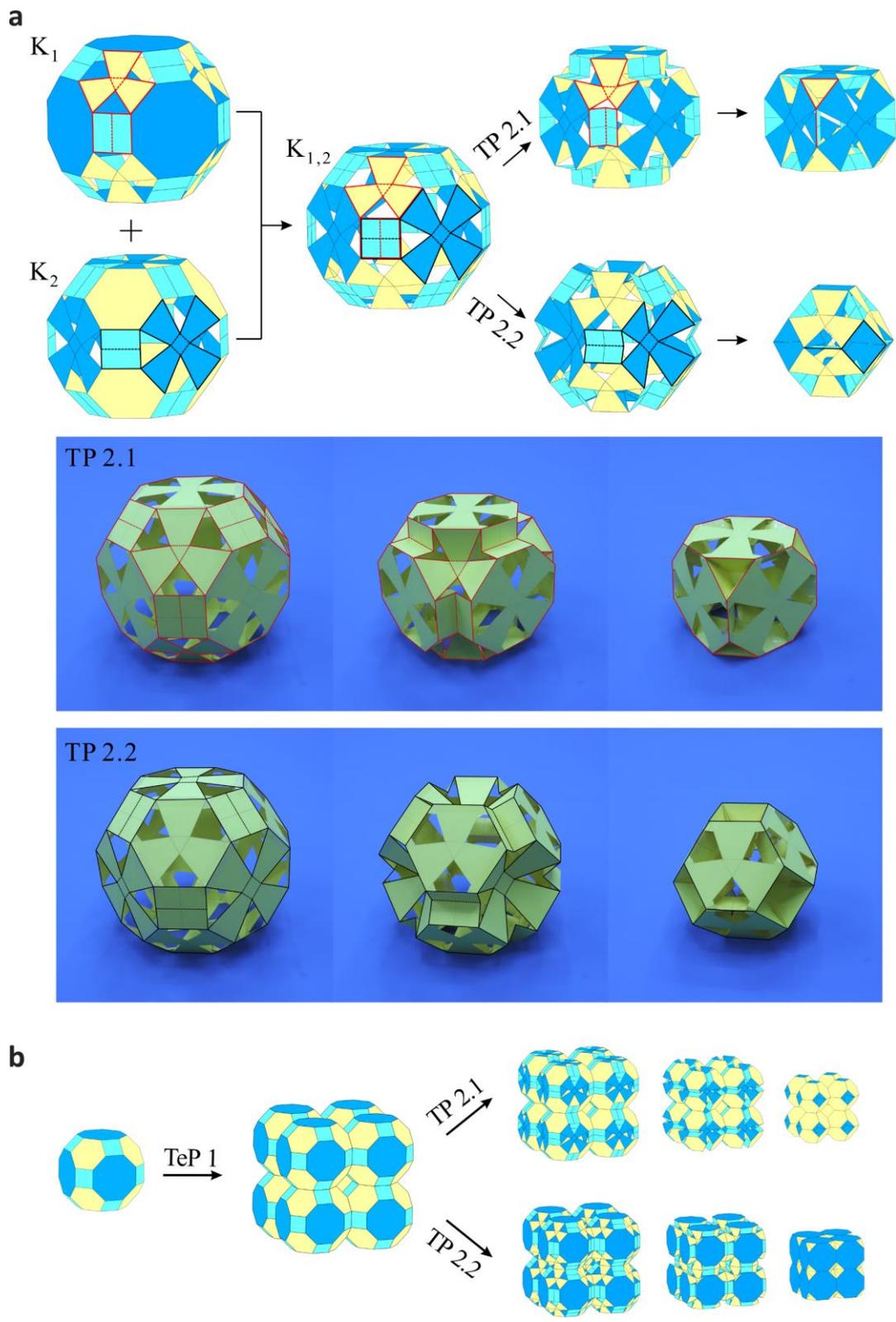

**Fig. 4 Superimposed pattern and polyhedral tessellation. a** The superimposed pattern $K_{1,2}$ with two different folding paths TP2.1 and TP2.2 (Supplementary Movie 4). **b** Polyhedral tessellations with $O_h$ symmetry and the folding motion following paths TP2.1 and TP2.2.